\documentclass[conference]{IEEEtran}
\usepackage{times}

\usepackage[numbers]{natbib}
\usepackage{amsmath}
\usepackage{amssymb}
\usepackage{multicol}
\usepackage[bookmarks=true]{hyperref}
\usepackage{graphicx}
\usepackage{subcaption}

\newcommand\blfootnote[1]{%
  \begingroup
  \renewcommand\thefootnote{}\footnote{#1}%
  \addtocounter{footnote}{-1}%
  \endgroup
}
\begin{document}

\title{Residual Policy Learning}

% You will get a Paper-ID when submitting a pdf file to the conference system
% \author{Author Names Omitted for Anonymous Review. Paper-ID [add your ID here]}

\author{\authorblockN{Tom Silver$^*$, Kelsey Allen$^*$, Josh Tenenbaum, Leslie Kaelbling}
\authorblockA{MIT\\
\{tslvr, krallen, jbt\}@mit.edu, lpk@csail.mit.edu}
}

%\author{\authorblockN{Michael Shell}
%\authorblockA{School of Electrical and\\Computer Engineering\\
%Georgia Institute of Technology\\
%Atlanta, Georgia 30332--0250\\
%Email: mshell@ece.gatech.edu}
%\and
%\authorblockA{Twentieth Century Fox\\
%\and
%\authorblockN{James Kirk\\ and Montgomery Scott}
%\authorblockA{Starfleet Academy\\
%San Francisco, California 96678-2391\\

% avoiding spaces at the end of the author lines is not a problem with
% conference papers because we don't use \thanks or \IEEEmembership

%Homer Simpson\authorrefmark{2},
%Montgomery Scott\authorrefmark{3} and
%Eldon Tyrell\authorrefmark{4}}
%Telephone: (800) 555--1212, Fax: (888) 555--1212}
%\authorblockA{\authorrefmark{4}Tyrell Inc., 123 Replicant Street, Los Angeles, California 90210--4321}}

\maketitle

\begin{abstract}
We present Residual Policy Learning (RPL): a simple method for improving nondifferentiable policies using model-free deep reinforcement learning. RPL thrives in complex robotic manipulation tasks where good but imperfect controllers are available. In these tasks, reinforcement learning from scratch remains data-inefficient or intractable, but learning a \textit{residual} on top of the initial controller can yield substantial improvements. We study RPL in six challenging MuJoCo tasks involving partial observability, sensor noise, model misspecification, and controller miscalibration. For initial controllers, we consider both hand-designed policies and model-predictive controllers with known or learned transition models. By combining learning with control algorithms, RPL can perform long-horizon, sparse-reward tasks for which reinforcement learning alone fails. Moreover, we find that RPL consistently and substantially improves on the initial controllers. We argue that RPL is a promising approach for combining the complementary strengths of deep reinforcement learning and robotic control, pushing the boundaries of what either can achieve independently\blfootnote{$^*$ Equal contribution.}\footnote{Video and code at \href{https://k-r-allen.github.io/residual-policy-learning/}{https://k-r-allen.github.io/residual-policy-learning/}.}.

\end{abstract}

\IEEEpeerreviewmaketitle

\section{Introduction}
\begin{figure}[t!]
\centering
  \includegraphics[width=0.5\textwidth]{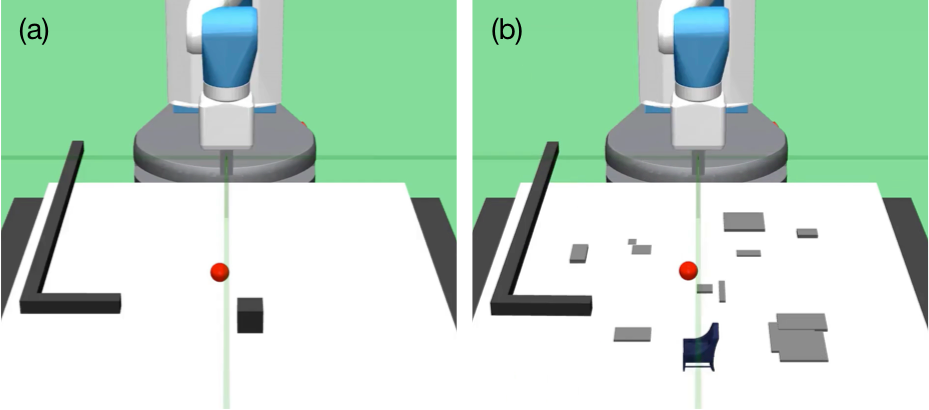}
  \caption{(a) A simulated Fetch robot must use a hook to move a block to a target (red sphere). A hand-designed policy can accomplish this task perfectly. (b) The same policy often fails in a more difficult task where the block is replaced by a complex object and the table top contains large ``bumps.'' Residual Policy Learning (RPL) augments the policy $\pi$ with a residual $f_{\theta}$, which can learn to accomplish the latter task.}
  \label{fig:hook-intro}
\end{figure}

Deep reinforcement learning (RL) methods are increasingly common and increasingly successful in robotic manipulation domains like grasping and pushing \cite{finn2017one, kalashnikov2018scalable, gu2017deep, zeng2018learning, her}. But for most complex problems of interest, learning from scratch remains intractable. For example, consider the task illustrated in Figure \ref{fig:hook-intro}a. A simulated Fetch robot must pick up and use a hook to drag an out-of-reach block to a target location. The only reward offered is a positive signal once the block reaches the target. This long-horizon, sparse-reward problem remains out of reach for current deep RL methods. In contrast, it is relatively straightforward to hand-design a policy that accomplishes this hook task perfectly in simulation (see Section \ref{sec:initial-policies}).

While a hand-designed policy may be robust to variations in the initial block position and target, it will likely break down with more dramatic variations in the task. For example, consider the task variation illustrated in Figure \ref{fig:hook-intro}b. The robot must now move a more complex rigid object to the goal. The task is further complicated by static ``bumps'' on the table that may impede the movement of the hook and object. Moreover, the robot's state includes no information about the bumps, which randomly regenerate at each trial, nor information about the object's shape, which is randomly selected from a library of 100 diverse objects. The policy designed for the original task sometimes succeeds in this setup, but more often fails.

What should be done when a policy --- be it a hand-designed policy, a model-predictive controller, or any other controller mapping states to actions --- performs below par? One path forward is to manually tweak the policy. This option, while potentially laborious, may suffice for some problems. But for other problems like the complex hook task described above, it is unclear how to even begin improving the policy by hand.

In this work, we propose Residual Policy Learning (RPL): a method for improving policies using deep reinforcement learning. Our main idea is to augment arbitrary initial policies by learning \textit{residuals} on top of them. Given an initial policy $\pi : \mathcal{S} \to \mathcal{A}$ with states $s \in \mathcal{S}$ and actions $a \in \mathcal{A} \subseteq \mathbb{R}^d$, we learn a residual function $f_\theta(s) : \mathcal{S} \to \mathcal{A}$ so that we have a \textit{residual policy} $\pi_\theta : \mathcal{S} \to \mathcal{A}$ given by
\begin{align*}
 \pi_\theta(s) = \pi(s) + f_\theta(s)\;\;.
 \label{residual-policy-eqn}
\end{align*}

Observe that $\nabla_{\theta} \pi_\theta(s) = \nabla_{\theta} f_\theta(s)$, that is, the gradient of the policy does not depend on the initial policy $\pi$. We can therefore use policy gradient methods to learn $\pi_\theta$ even if the initial policy $\pi$ is not differentiable.

There are two ways to see the role of the residual. If the initial policy is nearly perfect, the residual $f_\theta$ may be viewed as a corrective term. But if the initial policy is far from ideal, we may interpret the outputs of $\pi$ as merely ``hints'' to guide exploration. In practice, these two interpretations of the residual represent ends of a spectrum. We study problems all along this spectrum in this paper.

We present experimental results on several complex manipulation tasks that feature issues central to robotics and controller design: partial observability, sensor noise, model misspecification, and controller miscalibration. Our experiments are designed to investigate when and to what extent the following two claims hold:

\begin{enumerate}
    \item RPL improves on initial policies; and
    \item RPL is more data-efficient than learning from scratch. 
\end{enumerate}

\noindent We examine two common sources of initial policies: hand-designed policies and model-predictive controllers (MPC). We consider MPC with both known and learned transition models. In the latter case, we use Probabilistic Ensembles with Trajectory Sampling (PETS), a state-of-the-art method for model-based RL, to derive the initial controller \cite{chua2018deep}. In all cases, RPL is able to substantially improve on the original policies, while requiring far less data than learning from scratch to achieve the same performance. Furthermore, in complex manipulation tasks like that in Figure \ref{fig:hook-intro}b, RPL succeeds where learning from scratch is intractable and hand-designing perfect policies is unrealistic.

\section{Related Work}
RPL be seen as tackling two separate but related questions: how to improve imperfect controllers, and how to make deep reinforcement learning methods more data efficient and able to handle longer horizon planning.

There has been a substantial body of work on improving the data efficiency of deep RL by combining model-free and model-based approaches.
These methods often first learn a dynamics model and then use this dynamics model to simulate experience \cite{sutton1990integrated, deisenroth2015gaussian, gu2016continuous} or compute gradients for model-free updates \cite{heess2015learning, nguyen1990neural}.
Another set of approaches uses the learned dynamics model (or inverse dynamics model) to perform trajectory optimization or model-predictive control \cite{chua2018deep,mordatch2016combining}.
Further work uses such model-based methods to guide a model-free learner in a DAGGER-style imitation strategy \cite{nagabandi}.
More recent work has shown an equivalence between model-free and model-based RL with goal-conditioned value functions \cite{pong2018temporal}, and used this to improve model-free RL data efficiency.
RPL can be seen as an extension of this line of work, as it provides a new means for combining the benefits of model-based and model-free RL. We show in experiments that the model-based method proposed by \citet{chua2018deep} can be improved upon with RPL. However, RPL is also more general; it can be used to improve upon arbitrary policies, including but not limited to model-based ones.

RPL can also be seen as a form of imitation learning. 
This set of approaches considers an expert that provides demonstrations of a task to a learner.
Most approaches then attempt to copy the expert's strategy \cite{nagabandi, dagger}, or to use inverse reinforcement learning to infer goals and subgoals of the expert agent \cite{abbeel2004apprenticeship, ziebart2008maximum}.
Underpinning most of these approaches is the supposition that the expert is perfect. If the expert is indeed perfect, then RPL will be immediately perfect as well due to our initialization strategy (see Section \ref{sec:initializing}). But if the expert is imperfect and is only meant to provide ``hints,'' RPL learns to improve nonetheless.

From robotics, many methods exist for learning different aspects of the perception, control, execution pipeline.
Focusing on control specifically, Bayesian optimization approaches are popular for learning controllers based on Gaussian process models of objective functions to be optimized \cite{wang2018RSS, marco2017virtual, lizotte2007automatic,tesch2011using, marco2016automatic}.
Learning an accurate dynamics model is another central focus for robotics (termed system identification), and has been approached using analytic gradients \cite{wan2000unscented, de2018end}, finite differences \cite{kolev2015physically} or Bayesian Optimization \cite{deisenroth2015gaussian}.
In contrast, RPL does not presuppose which aspect of the controller needs correction.
This is particularly valuable in partially observable settings, where it is unclear how to learn a good dynamics model or design a better objective function.

In the case of dynamics learning, our work is inspired by \citet{ajay2018augmenting} and \citet{kloss2017combining} who learn a correction to an analytical physics model in order to perform better model-predictive control.
RPL is more general in that it can learn to correct the model implicitly by correcting the policy, but can also provide corrections which could not be provided by dynamics corrections (such as partially observable or noisy domains).

Concurrent work by \citet{concurrent-paper} also proposes residual reinforcement learning, and focuses on showing the value of the approach for real robots in a task of block insertion, investigating the effects of variation in the initial state, control noise, and the transfer from sim to real.
Here we aim to show the power of residual policies for a variety of different tasks that disentangle several sources of difficulty: partial observability, sensor noise, model misspecification, and controller miscalibration. We also empirically analyze the root cause of RPL's success by introducing a baseline that uses the initial policy only as an ``expert'' to guide exploration.

\section{Background}
\label{background}

\begin{figure*}[h!t]
  \includegraphics[width=\textwidth]{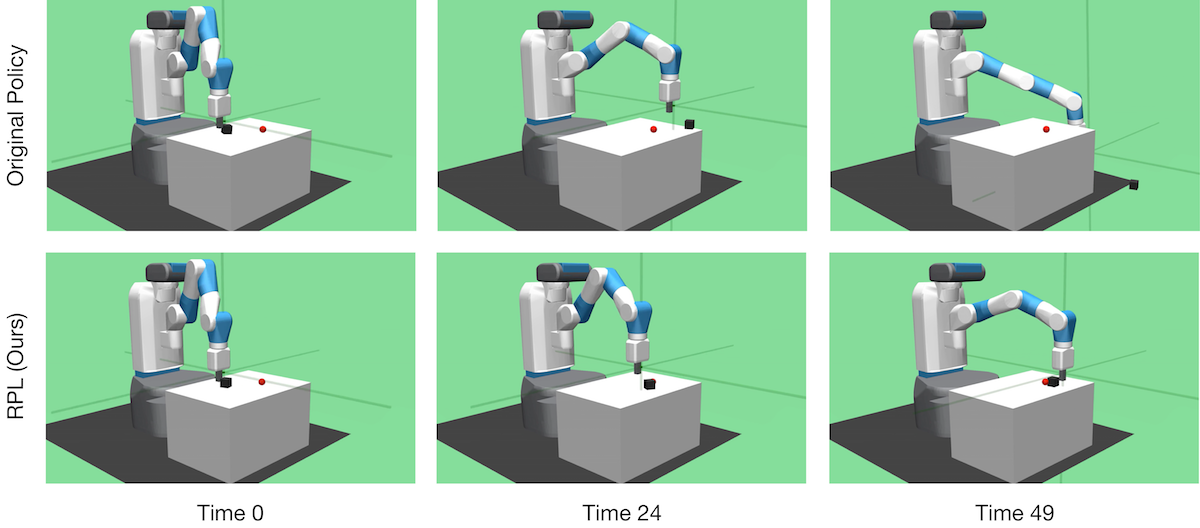}
  \caption{Illustration of the original \texttt{ReactivePush} policy and RPL on the \texttt{SlipperyPush} task. The original policy was designed so that the robot pushes the block to the target (red sphere) when the block has high friction. When the block has lower friction than anticipated, the block is pushed off the table (top row). RPL, our proposed method, learns to correct the faulty policy and accomplish the task after 1 million simulator steps (bottom row). Given the same number of time steps, reinforcement learning from scratch results in a policy where the robot does not touch the block at all (not shown).}
  \label{fig:slippery-push-sequence}
\end{figure*}

RPL operates within a standard (Partially Observable) Markov Decision Process (MDP) framework. An MDP is a tuple $M = (\mathcal{S}, \mathcal{A}, R, T, \gamma)$ where $s \in \mathcal{S}$ are states, $a \in \mathcal{A}$ are actions, $R(s, a) \in \mathbb{R}$ is the reward for taking action $a$ in state $s$, $T(s, a, s') = Pr(s' | s, a)$ is the probability of transitioning to state $s'$ following state $s$ and action $a$, and $0 \le \gamma \le 1$ is a temporal discount factor. We assume all trajectories or episodes sampled from the MDP have a finite number of actions (horizon) $h$. In all of the experiments described in this paper, states and actions are real-valued vectors. A policy $\pi : \mathcal{S} \to \mathcal{A}$ maps states to actions. Given an initial state $s_0$, the reinforcement learning problem is to find a policy $\pi$ that maximizes expected rewards discounted over time $J = \mathbb{E}_{s_t \sim M}[\sum_{t=0}^h \gamma^t R(s_t, \pi(s_t))]$. 

Let $Q^{\pi} : \mathcal{S} \times \mathcal{A} \to \mathbb{R}$ be the action-value function that gives the expected future discounted rewards following policy $\pi$. Many reinforcement learning methods make use of the Bellman equation for the action-value function $$Q^\pi(s, a) = \mathbb{E}_{s' \sim T(s, a, \cdot)} [R(s, a) + \gamma Q^{\pi}(s', \pi(s')) ]$$

\noindent Actor-critic methods learn both a parameterized policy $\pi_\theta$ (the actor) and a parameterized action-value function $Q_\theta$ (the critic). The critic is trained with a loss function derived from the Bellman equation above and the actor is trained to produce actions that maximize the critic. This approach is typically more stable than training the actor alone.

For the experiments in this work, we use Deep Deterministic Policy Gradients (DDPG) \cite{ddpg}, an actor-critic method that works well in domains with continuous states and actions (though any RL method could be used with RPL in principle). In DDPG, the actor is updated following the deterministic policy gradient $$ \nabla_{\theta} J \approx \mathbb{E}_{s_t \sim M}[\nabla_a Q_\theta(s, a) \rvert_{s=s_t, a = \pi_\theta(s_t)} \nabla_{\theta} \pi_\theta(s)\rvert_{s=s_t}] $$ DDPG makes use of experience replay, in which transitions $(s_t, a_t, r_t, s_{t+1})$ sampled from the environment are stored in a replay buffer. During training, transitions are then randomly drawn from the replay buffer in an effort to break the correlation between consecutive transitions.

Hindsight Experience Replay (HER) \cite{her} extends experience replay to dramatically improve data efficiency in domains with sparse binary rewards (goals) like those we consider in our experiments. In HER, the reward function and policy are additionally parameterized by a goal $g$ so that they become $R(s, a, g)$ and $\pi(s, g)$ respectively. For our purposes, the goal $g$ is a subvector of the final state of an episode. During training, each transition added to the replay buffer includes a goal $g_t$ that was achieved ``in hindsight,'' i.e. the goal that was actually reached at the end of the training episode. Given a sampled transition $(s_t, a_t, r_t, s_{t+1}, g_t)$, the policy is then updated according to the reward $R(s_t, a_t, g_t)$. This trick is especially useful early in training when the chance of achieving nonzero rewards is low. We combine HER and DDPG for all of the experiments presented in this work.

\section{Residual Policy Learning (RPL)}
\label{sec:rpl}

In Residual Policy Learning (RPL), we begin with an initial policy $\pi : \mathcal{S} \to \mathcal{A}$. Our goal is to learn a residual $f_\theta$ to create an improved final policy $\pi_\theta(s) = \pi(s) + f_\theta(s)$. 

Observe that a fixed initial policy $\pi$ together with an MDP $M= (\mathcal{S}, \mathcal{A}, R, T, \gamma)$ induces a \textit{residual MDP} $M^{(\pi)} = (\mathcal{S}, \mathcal{A}, R, T^{(\pi)}, \gamma)$ where 

$$T^{(\pi)}(s, a, s') = T(s, \pi(s) + a, s')$$

\noindent If we view $M^{(\pi)}$ as an MDP like any other, we see that the residual that we wish to learn, $f_\theta$, is a policy in this MDP. We can thus apply standard reinforcement learning techniques to learn the residual. In this work, we parameterize $f_\theta$ as a neural network and use model-free deep RL methods for learning.

RPL is as simple as that: given an initial policy, create a residual policy and proceed with deep RL. We now describe a few minor extensions that can improve performance and data efficiency in practice.

\subsection{Initializing the Residual}
\label{sec:initializing}
A desirable property of RPL is that it should never make a good initial policy worse. In the extreme case, if an initial policy is perfect, then we would like the residual policy to have no influence. We therefore endeavor to initialize the residual function so that $f_\theta(s) = \b{0}$ for all $s \in \mathcal{S}$. We do this by initializing the last layer of the network to be zero.

\subsection{RPL with Actor-Critic Methods}
\label{sec:warm-up-period}

RPL learns a residual on the output of an initial policy. Actor-critic methods like DDPG involve not only a policy but also a learned action-value function. If we begin with a perfect initial policy and a poor critic, the policy performance may degrade, since it is trained with reference to the critic. We therefore propose to train the critic alone for a ``burn in'' period while leaving the policy fixed. We can determine an appropriate burn in length automatically by monitoring the critic loss function and waiting for it to dip below a threshold $\beta$, which becomes a hyperparameter of our method. We use $\beta=1.0$ for all experiments in this paper.

\subsection{Recurrent RPL for POMDPs}
\label{sec:recurrent-rpl}
RPL can also be extended to handle Partially Observable Markov Decision Processes (POMDPs).
Generally, this is done in deep reinforcement learning by making $\pi_{\theta}(s)$ recurrent.
In practice, this is challenging for DDPG \cite{kapturowski2018recurrent}, and so we present an approximation by simply considering a "history" of previous states.
This is equivalent to writing $s = \{s_{t-n} \| n \in 0,...,N\}$ with $t$ being the current time-step, and $N$ the \emph{history length}. 
While the history length could take on any value, we found that a history length of just 1 (meaning the policy considers the current state and previous state) to be effective. We take advantage of this extension in our \texttt{NoisyHook} experiment in which observation noise obscures the input to the policy.

\section{Experiments}

Here we investigate to what extent RPL improves on initial policies, learns faster than model-free RL alone, and succeeds in tasks where model-free RL is intractable.

\subsection{Tasks}
\label{sec:tasks}

We study six simulated manipulation tasks. All environments are implemented in MuJoCo \cite{mujoco}. To provide direct comparison with previous work, we begin with a \texttt{Push} task and a \texttt{PickAndPlace} task, both taken from \citet{fetch}. We then present three more difficult tasks that have not been previously considered: \texttt{SlipperyPush}, \texttt{NoisyHook}, and \texttt{ComplexHook}. These first five tasks all involve a Fetch robot positioned in front of a table top. In the final task, we use the ``7-DOF Pusher'' environment from \citet{chua2018deep}. Our focus in the last task is model-based RL, so we call this environment \texttt{MBRLPusher}. 

In the first five tasks, following previous work, we parameterize the action space in terms of changes to the end effector's $xyz$ position in world coordinates \cite{fetch}. A fourth action coordinate modulates the gripper's two fingers symmetrically. (In the push tasks, the gripper is locked, and the fourth action coordinate has no effect.) In the sixth task, actions actuate the joints of the 7-DOF robot arm directly. All actions are normalized so that the resulting action space is $\mathcal{A} = [-1, 1]^D$, where $D=4$ for the first five tasks and $D=7$ for the last. The state spaces and rewards vary per task, as described next.

\subsubsection{\texttt{Push}}

This task is taken directly from \cite{fetch}. The objective is to push an object (a cube) to a target location on the table surface. The initial position of the object and the target location are randomized. The state space includes:

\begin{itemize}
    \item Gripper $xyz$ position and velocity (6 dims)
    \item Object $xyz$ position, $ypr$ rotation, velocities (12 dims)
    \item Object position relative to the gripper (3 dims)
    \item Gripper finger joint states and velocities (4 dims)
\end{itemize}

\noindent for a total dimensionality of 25. To use Hindsight Experience Replay, we must also specify achieved and desired goals. Here the achieved goal is the three-dimensional final position of the object and the desired goal is the target location. Rewards are sparse and binary: a reward of $1$ is given when the object is within a small radius around the target location and $0$ otherwise. The episode is counted as a success if the last reward is $1$, i.e. the goal is achieved. Episode lengths are 50 and do not terminate early.

\subsubsection{\texttt{SlipperyPush}}

Here we present a slight modification to the original \texttt{Push} environment. In the original environment, the object has a sliding friction coefficient of $1.0$. In this \texttt{SlipperyPush} environment, the same coefficient is set to $0.18$. The initial state randomization, state space, goals, rewards, and horizon are otherwise identical to \texttt{Push}.

\subsubsection{\texttt{PickAndPlace}}

This task is taken directly from \cite{fetch}. As in the previous tasks, the objective is to move an object (a cube) to a target location. However, the target location may now be either on the table top or in the air above the table. At the beginning of each episode, the $xy$ position for the target location is randomly sampled as before. Then with $0.5$ probability, the $z$ location is set to be on the table surface; otherwise, the $z$ location is randomly sampled to be above the table surface. As mentioned above, the gripper is now unlocked so that the fingers open and close following the fourth action dimension. All other environment details are unchanged with respect to \texttt{Push}.

\subsubsection{\texttt{NoisyHook}}

In this task, the robot cannot initially reach the block with its gripper. A new hook object is introduced and positioned to the right of the robot (see Figure \ref{fig:hook-intro}a). The objective is still to move the cube to a target location, but now the robot must use the hook to manipulate the cube. The target location is randomly initialized so that it lies between the cube and the robot. In addition to the 25 state dimensions included in the previous tasks, the state space now includes information about the hook:

\begin{itemize}
    \item Hook $xyz$ position, $rpy$ rotation, velocities (12 dims)
    \item Hook position relative to the gripper (3 dims)
\end{itemize}

\noindent for a total of $15+25 = 40$ dimensions. Rewards and goals are the same as in previous tasks; we provide no additional shaping rewards. 

This \texttt{NoisyHook} task is further complicated with the addition of observation noise. We suppose that the robot has precise proprioception but has significant uncertainty about the positions of the hook and cube. At each time step, we add IID diagonal Gaussian noise ($\mu=0.0, \sigma^2=0.025$) to the $xy$ position of the block and  the $xyz$ position of the hook, as well as the rotation of both objects. Since the achieved goals are derived from the state, they too are affected. Here we double the episode length for a total of 100 frames. 

\subsubsection{\texttt{ComplexHook}}

This task again features a hook and an object that must be moved to a target location. There is no longer noise added to the state. There is, however, significant uncertainty of two different, structured kinds. We first replace the simple cube from previous tasks with complex objects that vary in mass, friction, and shape. We use 100 objects taken from previous work by \citet{finn2017one}. The object meshes were originally downloaded from \href{http://thingiverse.com}{thingiverse.com} and include bowls, teddy bears, and small chairs among many other shapes. No information about the object shape or physical parameters is included in the state. To accomplish this task robustly, a policy must work across all possible objects.

To introduce a second source of structured uncertainty, we simulate large ``bumps'' on the table. A bump is a rigid box that is fixed to the table top. The width, length, height, position, and count of the bumps are randomly selected. See Figures \ref{fig:hook-intro}b and \ref{fig:complex-hook-sequence} for two examples. Note crucially that no information about the bumps are included in the state space. Thus the complete state space and other task parameters remain unchanged.

\subsubsection{\texttt{MBRLPusher}}

This final task is taken directly from \cite{chua2018deep}. A 7-DOF robot arm (not Fetch, but a simpler model) is positioned in front of a table with a tall cylinder and a target area. The objective is to push the cylinder to the target. The cylinder position and initial arm velocity are randomized per trial but the goal is fixed. The state space includes:
\begin{itemize}
    \item The robot joint positions and velocities ($14$ dims)
    \item The cylinder center of mass ($3$ dims)
    \item The gripper center of mass ($3$ dims)
\end{itemize}
for a total of 20 dimensions. Goals are the same as in previous tasks. Rewards are weighted sums of three terms: negative L1 norm between the cylinder and the goal, negative L1 norm between the gripper and the cylinder, and negative L2 norm of the action, with weights $(1.25, 0.1, 0.5)$ respectively. The task horizon here is 150 frames. 

\subsection{Initial Policies}
\label{sec:initial-policies}

In Residual Policy Learning (RPL), we begin with an environment and an initial policy $\pi : \mathcal{S} \to \mathcal{A}$ and we learn to improve on that initial policy. The initial policies that we use for our experiments are:
\begin{enumerate}
    \item \texttt{DiscreteMPCPush}: a model-predictive controller with discrete actions and heuristics specific to \texttt{Push}.
    \item \texttt{ReactivePush}: a reactive policy designed to work perfectly in the original \texttt{Push} task.
    \item \texttt{ReactivePickAndPlace}: a reactive policy for the \texttt{PickAndPlace} task with miscalibrated gains.
    \item \texttt{ReactiveHook}: a reactive policy designed to work perfectly in the noiseless hook task (Figure \ref{fig:hook-intro}a).
    \item \texttt{CachedPETS}: a model-predictive controller with a learned transition model \cite{chua2018deep}. To make this controller fast, we cache the output actions for 500 input states. Given a new state, the final \texttt{CachedPETS} controller finds the nearest state in the cache and outputs the corresponding stored action. The number 500 was selected based on a small performance analysis (see appendix). 
\end{enumerate}
See the appendix for details on all policies.

\subsection{Architectures and Training Details}
RPL is indifferent to the deep RL method applied or architecture used. However, for consistency, we use the same actor-critic architecture with Deep Deterministic Policy Gradients \cite{ddpg} and Hindsight Experience Replay \cite{her} across our experiments.
The network consists of 3 fully connected layers of 256 units each, with ReLU non-linearities (not on the output layer).
We use the same hyperparameters as in \cite{fetch}, given in the appendix. 
Our only substantial modification is to initialize the last layer of the network to zeros, so that the policy starts with the base controller (as described in section \ref{sec:initializing}).
%Thresholds (described in section \ref{sec:warm-up-period}) are also given in the appendix.

When training in noisy environments, we use a history of 1 (see Section \ref{sec:recurrent-rpl}). 
We considered two variants. 
In the first variant, the states are concatenated and fed to the network: $f_{\theta}(s_1, s_2)$.
In the second variant, we consider the average of the features obtained for the states: $0.5\left(f_{\theta}(s_1) + f_{\theta}(s_2)\right)$.
In practice, we found the second variant to work better, and so use it for all noisy environments.
 
\subsection{Baselines}
\label{sec:baselines}

We consider three baselines for all experiments. First, in all experiments, we show the result of running the initial policy without learning. Second, we show the result of learning from scratch with DDPG and HER. 

Our third baseline is designed to disentangle the causes of RPL's success. One hypothesis for why residual learning might be helpful is that the initial policy provides a smart means for exploration. The baseline, ``Expert Explore", uses the initial (``expert'') policy for exploration only. Actions are selected by selecting $z = rand(0, 1)$ and proceeding as follows:
\begin{align*}
    a &= \begin{cases}
    \pi(s) \text{  if $z < \epsilon\alpha$} \\
    rand() \text{  if $z < \epsilon(1 - \alpha)$} \\
    f_{\theta}(s) \text{  if $z > \epsilon$} \\
    \end{cases}
\end{align*}
where $\epsilon$ and $\alpha$ are hyperparameters that we selected with a small grid search. Thus the agent acts $(1 - \epsilon)\%$ of the time according to the learned policy, $(\epsilon \times \alpha) \%$ according to the expert, and the rest of the time takes random actions.
This baseline is similar to a policy-reuse method \cite{fernandez2006probabilistic}.

% $\epsilon$ and $\alpha$ were determined for our experiments by performing a small grid search to optimize performance for the hook environment. 
% These values are given in the appendix, and used for all experiments.
\subsection{Results}

Here we present empirical and qualitative results for RPL across the six complex manipulation tasks described in Section \ref{sec:tasks}. For each task, we show RPL's superior data efficiency and performance compared to the three baselines described in Section \ref{sec:baselines}. All empirical results are presented with mean and standard deviation across five random seeds.

\subsubsection{\texttt{DiscreteMPCPush} in \texttt{Push}}
\begin{figure*}[ht!]
    \centering
    \begin{subfigure}[t]{0.32\textwidth}
        \centering
        \includegraphics[width=\textwidth]{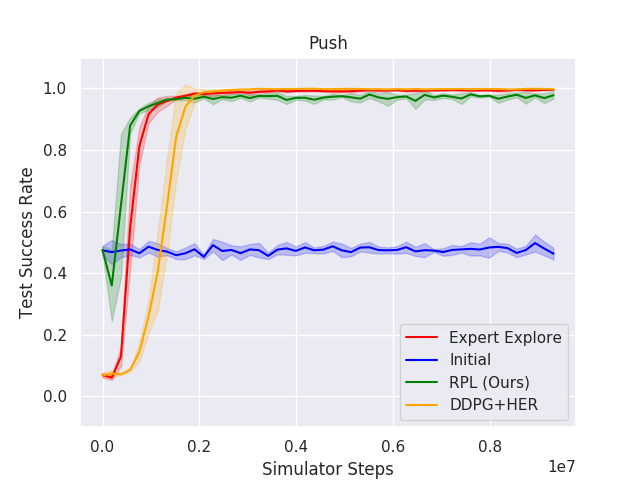}
    \end{subfigure}
    ~
    \begin{subfigure}[t]{0.32\textwidth}
        \centering
        \includegraphics[width=\textwidth]{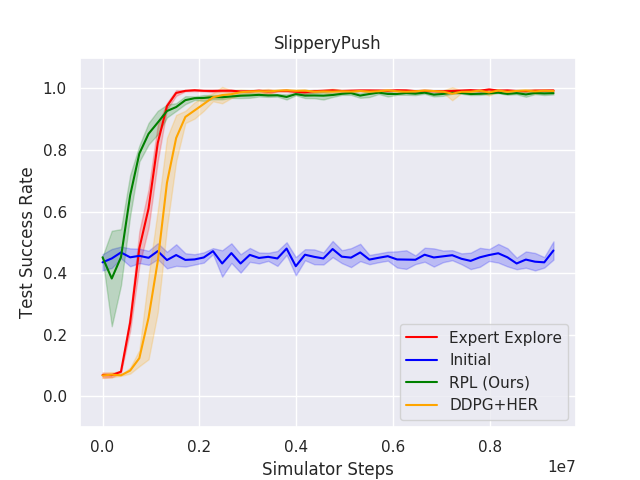}
    \end{subfigure}
    ~
    \begin{subfigure}[t]{0.32\textwidth}
        \centering
        \includegraphics[width=\textwidth]{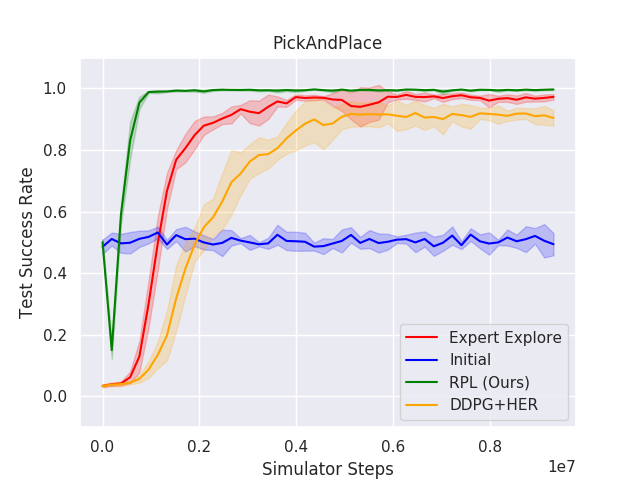}
    \end{subfigure}
    
    \caption{RPL and baseline results for the \texttt{Push}, \texttt{SlipperyPush}, and \texttt{PickAndPlace} tasks. As described in the text, the ``Initial'' policies are \texttt{DiscreteMPCPush}, \texttt{ReactivePush}, and \texttt{ReactivePickAndPlace} respectively. In the first two tasks, RPL converges to perfect performance in roughly the same number of simulation steps as learning from scratch with DDPG and HER but outperforms the baseline before convergence. In the third task, RPL converges with roughly 10x fewer training samples. In all cases, RPL substantially improves on the initial policies. }
    \label{fig:empirical-results-group1}
\end{figure*}
In this experiment, we examine whether RPL can overcome the limitations of an MPC controller that makes coarse approximations in an effort to trade performance for speed. In particular, we use the \texttt{DiscreteMPCPush} as our initial policy for the \texttt{Push} task.

We graph the success rates of RPL and the baselines in Figure \ref{fig:empirical-results-group1}a. The success rate of \texttt{DiscreteMPCPush} starts around 0.5. We noticed three common sources of suboptimality for this initial policy. First, the limited node expansions per MPC call, which is necessitated by the speed bottleneck of querying the MPC's model, means that a good action sequence is not always found. Second, the discreteness of the actions sometimes leads to circuitous executions in which the episode ends before the object reaches the target. Third, the heuristic used to guide the MPC's search, while very informative, can also be misleading in rare cases. 
These failure modes are especially common when the gripper must move from one side of the cube to the other, since the cube acts as an obstacle in this context.  

We confirm the results reported in previous work \cite{fetch} that learning from scratch with DDPG and HER works well in this domain, converging to a success rate of nearly 1.0 after roughly 2 million simulator steps. The performance of RPL before convergence greatly surpasses both the initial policy and learning from scratch, while still converging to a perfect success rate. For example, RPL takes an order of magnitude fewer training samples to reach an average success rate of 0.9 versus the learning from scratch baseline. 

Note that the performance of RPL drops early in training before quickly recovering and surpassing the baselines. We see this pattern in the following experiments as well. This is a manifestation of the issue discussed in Section \ref{sec:warm-up-period} whereby the critic is initialized poorly with respect to the actor. We found that decreasing the burn-in parameter $\beta$ mitigated the drop but did not significantly affect the time to convergence. We thus left the results as they are for the benefit of discussion.

To analyze the source of RPL's superior data efficiency, we turn to the performance of the Expert Explore baseline. We find that this baseline also improves on learning from scratch, but that RPL converges slightly faster. This suggests that RPL's advantage in this \texttt{Push} task derives in large part from more efficient exploration, but also from the residual parameterization and initialization.

\subsubsection{\texttt{ReactivePush} in \texttt{SlipperyPush}}

Our second experiment examines model misspecification. We tuned the \texttt{ReactivePush} policy to achieve near perfect performance in the \textit{original} \texttt{Push} task. We now transfer this policy to the \texttt{SlipperyPush} task in which the sliding friction coefficient of the cube is 5x smaller.

The success rates of RPL and the baselines on the \texttt{SlipperyPush} task are shown in Figure \ref{fig:empirical-results-group1}b. As expected, the \texttt{ReactivePush} policy is not perfect, achieving a success rate of around 0.45. The most common failure mode of this initial policy is when the gripper pushes the slippery cube too hard and the cube slides off the table. In other cases, the cube does not fall off, but is pushed back and forth across the goal without converging. A representative trial is illustrated in Figure \ref{fig:slippery-push-sequence} (top row). As in the first experiment, we find that RPL is far better before convergence and converges to the same perfect success rate as model-free learning from scratch.

\subsubsection{\texttt{ReactivePickAndPlace} in \texttt{PickAndPlace}}

In this experiment, we consider an example of a poorly calibrated initial policy that leads to detrimental oscillatory behavior. Such oscillations are a common issue in stateless robotic control when gains are improperly tuned. To create a representative scenario, we start with the \texttt{ReactivePickAndPlace} policy and artificially increase the gains. Oscillations quickly arise, e.g. when the gripper overshoots the waypoints implicit in the design of the policy. These oscillations cause the success rate of the \texttt{ReactivePickAndPlace} to drop to roughly 0.5, as seen in Figure \ref{fig:empirical-results-group1}c.

As reported in previous work \cite{fetch}, learning from scratch with DDPG and HER requires far more data to reach a success rate of 1.0 in \texttt{PickAndPlace} versus \texttt{Push}. Here we find the data efficiency of RPL to be substantially better. RPL converges to a success rate of 1.0 after roughly 1 million simulator steps, which represents a nearly 10x improvement over learning from scratch. Comparing with the Expert Explore baseline, we find that not all of the advantage can be explained by improved exploration; the good parameterization and initialization of the policy is also to credit.

It was not a priori obvious that the initial policy would aid RPL here as much as it apparently does. By design, we know that the policy is close in ``gain space'' to a near optimal one, but that does not guarantee that the policy is similarly close in ``residual weight space.'' Fortunately, it seems the two notions coincide here. 

%A second interesting observation is that the performance of RPL drops starkly early in training before quickly recovering and surpassing the baselines. This is a manifestation of the issue discussed in Section \ref{sec:warm-up-period} whereby the critic is initialized poorly with respect to the actor. We found that decreasing the burn-in parameter $\beta$ mitigated the drop here but did not significantly affect the time to convergence. We thus left the results as they are for the benefit of discussion.

\begin{figure*}[ht!]
    \centering
    \begin{subfigure}[t]{0.32\textwidth}
        \centering
        \includegraphics[width=\textwidth]{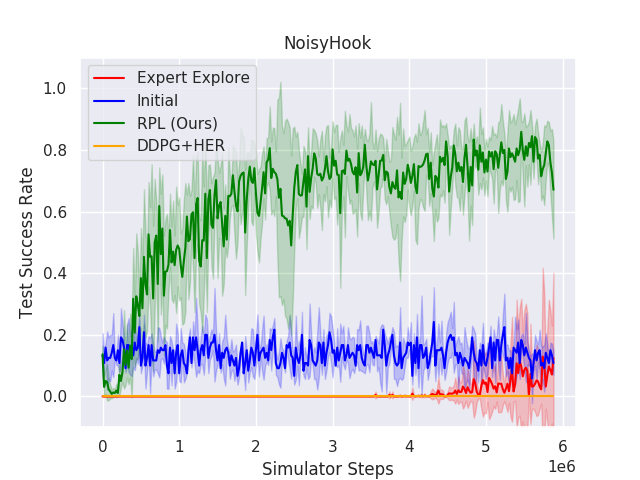}
    \end{subfigure}
    ~
    \begin{subfigure}[t]{0.32\textwidth}
        \centering
        \includegraphics[width=\textwidth]{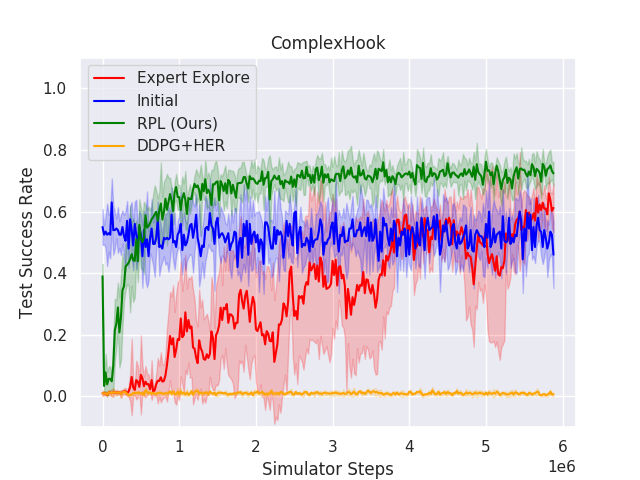}
    \end{subfigure}
    ~
    \begin{subfigure}[t]{0.32\textwidth}
        \centering
        \includegraphics[width=\textwidth]{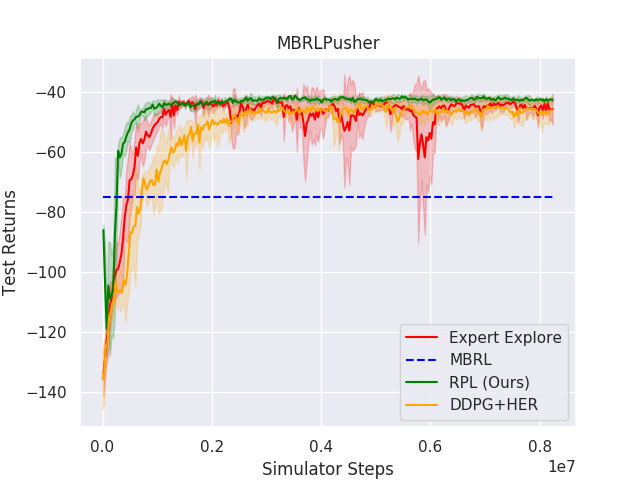}
    \end{subfigure}
    \caption{RPL and baseline results for the \texttt{NoisyPush}, \texttt{ComplexPush} and \texttt{MBRLPusher} tasks. For the first two tasks, the ``Initial'' policy is \texttt{ReactiveHook}; for the third, it is \texttt{CachedPETS}. In the first task, RPL quickly and substantially improves on the initial policy while the other two learning methods fail. In the second task, RPL again improves on the initial policy. See also Figures \ref{fig:hook-intro}b and \ref{fig:complex-hook-sequence} for illustrations of this task. In the third, RPL improves on the initial policy, converges faster than DDPG+HER, and outperforms the model-based reinforcement learning method PETS (``MBRL''). The dashed line marks the average performance of PETS over 10 trials.} .
    \label{fig:empirical-results-group2}
\end{figure*}

\begin{figure*}[ht!]
  \includegraphics[width=\textwidth]{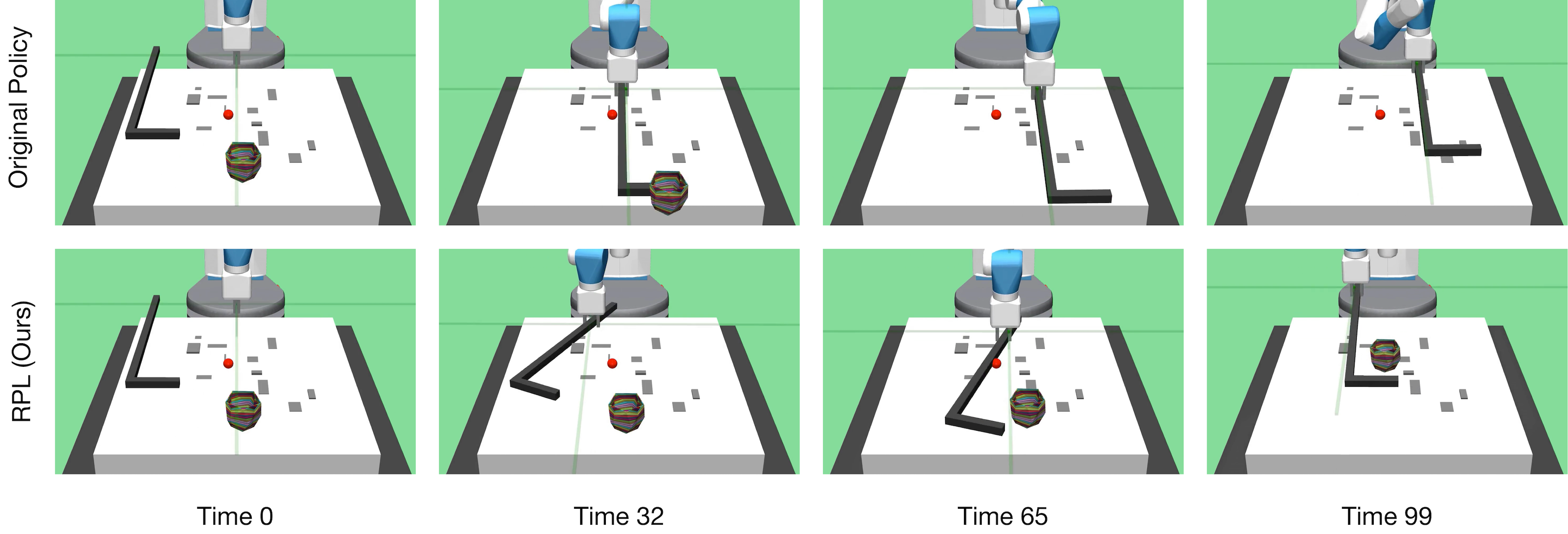}
  \caption{Illustration of the original policy and RPL on the \texttt{ComplexHook} task. The original policy was designed to work when the object is a simple cube and the table has no obstacles (see Figure \ref{fig:hook-intro}a). The same policy pushes a larger complex object off the table rather than to the target (red sphere) as required (top row). RPL, our proposed method, learns to improve the policy that pulls the object to the target (bottom row). The learned policy exhibits interesting behavior that qualitatively resembles lifting the hook to avoid obstacles and reaching around at a wider angle than originally programmed.}
  \label{fig:complex-hook-sequence}
\end{figure*}
\subsubsection{\texttt{ReactiveHook} in \texttt{NoisyHook}}

Now we turn to another prevalent problem in robotic control --- sensor noise --- and investigate whether RPL can improve the robustness of a sensitive initial policy. As discussed in Section \ref{sec:tasks}, the \texttt{NoisyHook} task features Gaussian noise applied to the positions and rotations of the block and hook. While the \texttt{ReactiveHook} policy is nearly perfect in a noiseless version of the same task, the policy proves to be quite sensitive to the sensor noise. We observe diverse failure modes throughout the course of execution: the gripper often moves to a wrong position, sometimes fails to pick up the hook, and other times drops the hook. As shown in Figure \ref{fig:empirical-results-group2}a, the success rate of the initial policy is roughly $0.15$, far lower than in our previous experiments.

In this experiment, we make use of the two frame policy architecture described in Section \ref{sec:recurrent-rpl} to cope with sensor noise. We use the same architecture for all three learning methods for comparison.

Learning from scratch with DDPG and HER fails in this task, never achieving a nontrivial success rate. This failure is not surprising given the long horizon and sparse rewards in the task. The Expert Explore baseline also performs quite poorly, only beginning to reach nontrivial success rates after 5 million simulator steps. We speculate that this failure is due to the fact that the hook is so often dropped by the initial policy. 

In contrast, we see that RPL quickly converges to a success rate of roughly 0.8. This represents the first instance of RPL obtaining strong performance in a task that is both out of reach for current deep RL methods and nontrivial for robotic control alone. Moreover, the results suggest that RPL is a promising method for overcoming the common challenge of sensor noise.

\subsubsection{\texttt{ReactiveHook} in \texttt{ComplexHook}}

In this experiment, we study structured uncertainty inspired by the common mismatch between physics simulators and real robotics tasks. As described in Section \ref{sec:tasks}, the \texttt{ComplexHook} task contains two challenging innovations over the noiseless hook task: bumps are randomly scattered across the table surface; and the object takes on a variety of shapes, masses, and coefficients of friction. We observed that each of these two innovations independently cause the \texttt{ReactiveHook} policy performance to drop by roughly 20\%. With both changes present, the initial policy success rate drops to 0.55, as shown in Figure \ref{fig:empirical-results-group2}b.

A random or null policy is occasionally successful in this task due to the scene randomization. With this in mind, we see that learning from scratch with DDPG and HER does not obtain any nontrivial success rate, as in the previous experiment. We again find that the policy never causes the gripper to touch the hook, let alone move it to reach the object.

Interestingly, the Expert Explore baseline does achieve a nontrivial success rate, eventually slightly surpassing the success rate of the initial policy. This task is easier than \texttt{NoisyHook} from the perspective of the expert baseline if only because the initial success rate is much higher.

Finally, RPL learns a robust policy with strong data efficiency, converging at a success rate just below 0.8. The fact that RPL is able to achieve this success rate is fairly remarkable given the diversity in the objects and obstacles, and the fact that the state contains no information about this diversity. RPL has apparently learned a ``conformant'' policy that works for most objects and obstacles without discretion. We show one intriguing example of RPL succeeding where the initial policy fails in Figure \ref{fig:complex-hook-sequence}.

\subsubsection{\texttt{CachedPETS} in \texttt{MBRLPusher}}

In this final experiment, we examine whether RPL can improve on a model-based RL method while converging faster than model-free RL. As described in Section \ref{sec:tasks}, to derive the initial controller, we begin by learning a transition model. Following \citet{chua2018deep}, we train for 15000 simulator steps in their \texttt{MBRLPusher} environment. We then query the environment for an additional 500 steps to construct our \texttt{CachedPETS} controller. (We thus offset our plotted results by 15500 relative to the baseline in Figure \ref{fig:empirical-results-group2}c.) The performance of PETS, averaged over 10 trials, is plotted as a dashed line in Figure \ref{fig:empirical-results-group2}c. We see that the drop in performance due to the caching approximation is fairly small.

We find that RPL improves substantially not only on the initial \texttt{CachedPETS} controller, but also on the original PETS controller. Furthermore, RPL converges faster than DDPG+HER, indicating that the initial controller was beneficial. It is worth emphasizing that no domain knowledge was used to design the initial policy here; this same combination of MBRL and RPL could be applied immediately to a new domain. These results suggest that RPL may be seen as a general RL method that marries the data efficiency of MBRL with the superior asymptotic performance of model-free RL.

\section{Discussion and Conclusion}

We have described Residual Policy Learning (RPL), a simple method that combines the strengths of deep RL and robotic control. Our experimental results suggest that RPL is a powerful approach to deal with pervasive issues in complex manipulation tasks such as sensor noise, model misspecification, controller miscalibration, and partial observability. We find that RPL consistently improves on initial policies and achieves better data efficiency than model-free RL alone. Furthermore, RPL can improve on initial policies for long-horizon, sparse-reward problems where model-free RL fails.

We have also seen the promise of combining RPL with model-based RL \cite{chua2018deep}. MBRL is often more data-efficient whereas model-free RL can be faster to run and asymptotically superior. RPL offers a simple mechanism for combining the strengths of both. We find empirically that learning a residual on top of MBRL improves on MBRL alone, converging to the same performance as model-free RL with less data.

We postulate three main causes for the success of RPL. First, as described in Section \ref{sec:rpl}, we take care to initialize the residual policy so that its output at first matches the initial policy. When the initial policy is strong, this initialization gives RPL a clear boost. The second cause of RPL's success is improved exploration early on during training. In learning from scratch with sparse rewards and long horizons, the first successful trajectory must be discovered by chance. Hindsight Experience Replay is designed to face this challenge, but RPL offers a more direct solution. RPL can discover successful trajectories immediately if the initial policy produces them with nontrivial frequency. To measure the impact of this exploration advantage, we introduced the Expert Explore baseline described in Section \ref{sec:baselines}. Empirically we find this baseline performance to lie midway between RPL and learning from scratch. A third likely cause of RPL's success is that the residual reinforcement learning problem induced by the initial policy may be easier than the original problem. This cause may best explain the superior performance of RPL in the \texttt{NoisyHook} task, where both the initial policy and the ``Expert Explore'' baseline are empirically poor.

% Further, RPL is able to improve upon hard-coded policies in part because of its ability to "compress" state representations for action. 
% We believe this is particularly helpful for handling noisy sensor data.

Though the six case studies we have presented all involve robotic manipulation with DDPG and HER, RPL is far more general than any specific task domain or deep RL method. The method we have described can be immediately applied in any domain with continuous actions and with any gradient-based learning method. However, RPL is especially well suited for complex manipulation because of the availability of good but imperfect initial policies and the long-horizon, sparse-reward tasks that naturally arise.

In recent years, complex manipulation problems have been at the forefront of research in robotics and deep RL. Both fields have made significant strides in often complementary directions. RPL should be viewed as one piece of a larger effort to combine the strengths of both approaches. We conjecture that solving the hardest open problems in manipulation will require such a synthesis.

% \newpage
% \newpage

\section*{Acknowledgments}
We thank Evan Shelhamer for helpful discussions. We gratefully acknowledge support from NSF grants 1523767 and 1723381; from ONR grant N00014-13-1-0333; from AFOSR grant FA9550-17-1-0165; from ONR grant N00014-18-1-2847; from Honda Research; and from the Center for Brains, Minds and Machines (CBMM), funded
by NSF STC award CCF-1231216. KA acknowledges support from NSERC. Any opinions, findings, and conclusions or recommendations expressed in this material are those of the authors and do not necessarily reflect the views of our sponsors.

\bibliographystyle{unsrtnat}
\bibliography{references}

\begin{thebibliography}{34}
\providecommand{\natexlab}[1]{#1}
\providecommand{\url}[1]{\texttt{#1}}
\expandafter\ifx\csname urlstyle\endcsname\relax
  \providecommand{\doi}[1]{doi: #1}\else
  \providecommand{\doi}{doi: \begingroup \urlstyle{rm}\Url}\fi

\bibitem[Finn et~al.(2017)Finn, Yu, Zhang, Abbeel, and Levine]{finn2017one}
Chelsea Finn, Tianhe Yu, Tianhao Zhang, Pieter Abbeel, and Sergey Levine.
\newblock One-shot visual imitation learning via meta-learning.
\newblock In \emph{Conference on Robot Learning}, pages 357--368, 2017.

\bibitem[Kalashnikov et~al.(2018)Kalashnikov, Irpan, Pastor, Ibarz, Herzog,
  Jang, Quillen, Holly, Kalakrishnan, Vanhoucke,
  et~al.]{kalashnikov2018scalable}
Dmitry Kalashnikov, Alex Irpan, Peter Pastor, Julian Ibarz, Alexander Herzog,
  Eric Jang, Deirdre Quillen, Ethan Holly, Mrinal Kalakrishnan, Vincent
  Vanhoucke, et~al.
\newblock Scalable deep reinforcement learning for vision-based robotic
  manipulation.
\newblock In \emph{Conference on Robot Learning}, pages 651--673, 2018.

\bibitem[Gu et~al.(2017)Gu, Holly, Lillicrap, and Levine]{gu2017deep}
Shixiang Gu, Ethan Holly, Timothy Lillicrap, and Sergey Levine.
\newblock Deep reinforcement learning for robotic manipulation with
  asynchronous off-policy updates.
\newblock In \emph{IEEE International Conference on Robotics and Automation
  (ICRA)}, pages 3389--3396. IEEE, 2017.

\bibitem[Zeng et~al.(2018)Zeng, Song, Welker, Lee, Rodriguez, and
  Funkhouser]{zeng2018learning}
Andy Zeng, Shuran Song, Stefan Welker, Johnny Lee, Alberto Rodriguez, and
  Thomas Funkhouser.
\newblock Learning synergies between pushing and grasping with self-supervised
  deep reinforcement learning.
\newblock \emph{arXiv preprint arXiv:1803.09956}, 2018.

\bibitem[Andrychowicz et~al.(2017)Andrychowicz, Wolski, Ray, Schneider, Fong,
  Welinder, McGrew, Tobin, Abbeel, and Zaremba]{her}
Marcin Andrychowicz, Filip Wolski, Alex Ray, Jonas Schneider, Rachel Fong,
  Peter Welinder, Bob McGrew, Josh Tobin, OpenAI~Pieter Abbeel, and Wojciech
  Zaremba.
\newblock Hindsight experience replay.
\newblock In \emph{Advances in Neural Information Processing Systems}, pages
  5048--5058, 2017.

\bibitem[Chua et~al.(2018)Chua, Calandra, McAllister, and Levine]{chua2018deep}
Kurtland Chua, Roberto Calandra, Rowan McAllister, and Sergey Levine.
\newblock Deep reinforcement learning in a handful of trials using
  probabilistic dynamics models.
\newblock \emph{arXiv preprint arXiv:1805.12114}, 2018.

\bibitem[Sutton(1990)]{sutton1990integrated}
Richard~S Sutton.
\newblock Integrated architectures for learning, planning, and reacting based
  on approximating dynamic programming.
\newblock In \emph{Machine Learning Proceedings 1990}, pages 216--224.
  Elsevier, 1990.

\bibitem[Deisenroth et~al.(2015)Deisenroth, Fox, and
  Rasmussen]{deisenroth2015gaussian}
Marc~Peter Deisenroth, Dieter Fox, and Carl~Edward Rasmussen.
\newblock Gaussian processes for data-efficient learning in robotics and
  control.
\newblock \emph{IEEE Transactions on Pattern Analysis and Machine
  Intelligence}, 37\penalty0 (2):\penalty0 408--423, 2015.

\bibitem[Gu et~al.(2016)Gu, Lillicrap, Sutskever, and Levine]{gu2016continuous}
Shixiang Gu, Timothy Lillicrap, Ilya Sutskever, and Sergey Levine.
\newblock Continuous deep {Q}-learning with model-based acceleration.
\newblock In \emph{International Conference on Machine Learning}, pages
  2829--2838, 2016.

\bibitem[Heess et~al.(2015)Heess, Wayne, Silver, Lillicrap, Erez, and
  Tassa]{heess2015learning}
Nicolas Heess, Gregory Wayne, David Silver, Timothy Lillicrap, Tom Erez, and
  Yuval Tassa.
\newblock Learning continuous control policies by stochastic value gradients.
\newblock In \emph{Advances in Neural Information Processing Systems}, pages
  2944--2952, 2015.

\bibitem[Nguyen and Widrow(1990)]{nguyen1990neural}
Derrick~H Nguyen and Bernard Widrow.
\newblock Neural networks for self-learning control systems.
\newblock \emph{IEEE Control Systems Magazine}, 10\penalty0 (3):\penalty0
  18--23, 1990.

\bibitem[Mordatch et~al.(2016)Mordatch, Mishra, Eppner, and
  Abbeel]{mordatch2016combining}
Igor Mordatch, Nikhil Mishra, Clemens Eppner, and Pieter Abbeel.
\newblock Combining model-based policy search with online model learning for
  control of physical humanoids.
\newblock In \emph{IEEE International Conference on Robotics and Automation
  (ICRA)}, pages 242--248. IEEE, 2016.

\bibitem[Nagabandi et~al.(2018)Nagabandi, Kahn, Fearing, and Levine]{nagabandi}
Anusha Nagabandi, Gregory Kahn, Ronald~S Fearing, and Sergey Levine.
\newblock Neural network dynamics for model-based deep reinforcement learning
  with model-free fine-tuning.
\newblock In \emph{IEEE International Conference on Robotics and Automation
  (ICRA)}, pages 7559--7566. IEEE, 2018.

\bibitem[Pong et~al.(2018)Pong, Gu, Dalal, and Levine]{pong2018temporal}
Vitchyr Pong, Shixiang Gu, Murtaza Dalal, and Sergey Levine.
\newblock Temporal difference models: Model-free deep rl for model-based
  control.
\newblock \emph{International Conference on Learning Representations}, 2018.

\bibitem[Ross et~al.(2011)Ross, Gordon, and Bagnell]{dagger}
St{\'e}phane Ross, Geoffrey Gordon, and Drew Bagnell.
\newblock A reduction of imitation learning and structured prediction to
  no-regret online learning.
\newblock In \emph{Proceedings of the Fourteenth International Conference on
  Artificial Intelligence and Statistics}, pages 627--635, 2011.

\bibitem[Abbeel and Ng(2004)]{abbeel2004apprenticeship}
Pieter Abbeel and Andrew~Y Ng.
\newblock Apprenticeship learning via inverse reinforcement learning.
\newblock In \emph{Proceedings of the Twenty-first International Conference on
  Machine Learning}, page~1. ACM, 2004.

\bibitem[Ziebart et~al.(2008)Ziebart, Maas, Bagnell, and
  Dey]{ziebart2008maximum}
Brian~D Ziebart, Andrew~L Maas, J~Andrew Bagnell, and Anind~K Dey.
\newblock Maximum entropy inverse reinforcement learning.
\newblock In \emph{AAAI Conference on Artificial Intelligence}, volume~8, pages
  1433--1438. Chicago, IL, USA, 2008.

\bibitem[Wang et~al.(2018)Wang, Garrett, Kaelbling, and
  Lozano{-}P{\'{e}}rez]{wang2018RSS}
Zi~Wang, Caelan~Reed Garrett, Leslie~Pack Kaelbling, and Tom{\'{a}}s
  Lozano{-}P{\'{e}}rez.
\newblock Active model learning and diverse action sampling for task and motion
  planning.
\newblock \emph{IEEE/RSJ International Conference on Intelligent Robots and
  Systems (IROS)}, 2018.

\bibitem[Marco et~al.(2017)Marco, Berkenkamp, Hennig, Schoellig, Krause,
  Schaal, and Trimpe]{marco2017virtual}
Alonso Marco, Felix Berkenkamp, Philipp Hennig, Angela~P Schoellig, Andreas
  Krause, Stefan Schaal, and Sebastian Trimpe.
\newblock Virtual vs. real: Trading off simulations and physical experiments in
  reinforcement learning with bayesian optimization.
\newblock In \emph{IEEE International Conference on Robotics and Automation
  (ICRA)}, pages 1557--1563. IEEE, 2017.

\bibitem[Lizotte et~al.(2007)Lizotte, Wang, Bowling, and
  Schuurmans]{lizotte2007automatic}
Daniel~J Lizotte, Tao Wang, Michael~H Bowling, and Dale Schuurmans.
\newblock Automatic gait optimization with {G}aussian process regression.
\newblock In \emph{International Joint Conference on Artificial Intelligence
  (IJCAI)}, pages 944--949, 2007.

\bibitem[Tesch et~al.(2011)Tesch, Schneider, and Choset]{tesch2011using}
Matthew Tesch, Jeff Schneider, and Howie Choset.
\newblock Using response surfaces and expected improvement to optimize snake
  robot gait parameters.
\newblock In \emph{IEEE/RSJ International Conference on Intelligent Robots and
  Systems (IROS)}, pages 1069--1074. IEEE, 2011.

\bibitem[Marco et~al.(2016)Marco, Hennig, Bohg, Schaal, and
  Trimpe]{marco2016automatic}
Alonso Marco, Philipp Hennig, Jeannette Bohg, Stefan Schaal, and Sebastian
  Trimpe.
\newblock Automatic {LQR} tuning based on {G}aussian process global
  optimization.
\newblock In \emph{IEEE International Conference on Robotics and Automation
  (ICRA)}, pages 270--277. IEEE, 2016.

\bibitem[Wan and Van Der~Merwe(2000)]{wan2000unscented}
Eric~A Wan and Rudolph Van Der~Merwe.
\newblock The unscented kalman filter for nonlinear estimation.
\newblock In \emph{Adaptive Systems for Signal Processing, Communications, and
  Control Symposium}, pages 153--158. Ieee, 2000.

\bibitem[de~Avila Belbute-Peres et~al.(2018)de~Avila Belbute-Peres, Smith,
  Allen, Tenenbaum, and Kolter]{de2018end}
Filipe de~Avila Belbute-Peres, Kevin Smith, Kelsey Allen, Josh Tenenbaum, and
  J~Zico Kolter.
\newblock End-to-end differentiable physics for learning and control.
\newblock In \emph{Advances in Neural Information Processing Systems}, pages
  7176--7187, 2018.

\bibitem[Kolev and Todorov(2015)]{kolev2015physically}
Svetoslav Kolev and Emanuel Todorov.
\newblock Physically consistent state estimation and system identification for
  contacts.
\newblock In \emph{IEEE-RAS 15th International Conference on Humanoid Robots
  (Humanoids)}, pages 1036--1043. IEEE, 2015.

\bibitem[Ajay et~al.(2018)Ajay, Wu, Fazeli, Bauza, Kaelbling, Tenenbaum, and
  Rodriguez]{ajay2018augmenting}
Anurag Ajay, Jiajun Wu, Nima Fazeli, Maria Bauza, Leslie~P. Kaelbling,
  Joshua~B. Tenenbaum, and Alberto Rodriguez.
\newblock Augmenting physical simulators with stochastic neural networks: Case
  study of planar pushing and bouncing.
\newblock \emph{arXiv preprint arXiv:1808.03246}, 2018.

\bibitem[Kloss et~al.(2017)Kloss, Schaal, and Bohg]{kloss2017combining}
Alina Kloss, Stefan Schaal, and Jeannette Bohg.
\newblock Combining learned and analytical models for predicting action
  effects.
\newblock \emph{arXiv preprint arXiv:1710.04102}, 2017.

\bibitem[Johannink et~al.(2018)Johannink, Bahl, Nair, Luo, Kumar, Loskyll,
  Ojea, Solowjow, and Levine]{concurrent-paper}
Tobias Johannink, Shikhar Bahl, Ashvin Nair, Jianlan Luo, Avinash Kumar,
  Matthias Loskyll, Juan~Aparicio Ojea, Eugen Solowjow, and Sergey Levine.
\newblock Residual reinforcement learning for robot control.
\newblock \emph{arXiv preprint arxIV:1812.03201}, 2018.

\bibitem[Lillicrap et~al.(2016)Lillicrap, Hunt, Pritzel, Heess, Erez, Tassa,
  Silver, and Wierstra]{ddpg}
Timothy~P Lillicrap, Jonathan~J Hunt, Alexander Pritzel, Nicolas Heess, Tom
  Erez, Yuval Tassa, David Silver, and Daan Wierstra.
\newblock Continuous control with deep reinforcement learning.
\newblock 2016.

\bibitem[Kapturowski et~al.(2019)Kapturowski, Ostrovski, Dabney, Quan, and
  Munos]{kapturowski2018recurrent}
Steven Kapturowski, Georg Ostrovski, Will Dabney, John Quan, and Remi Munos.
\newblock Recurrent experience replay in distributed reinforcement learning.
\newblock In \emph{International Conference on Learning Representations}, 2019.
\newblock URL \url{https://openreview.net/forum?id=r1lyTjAqYX}.

\bibitem[Todorov et~al.(2012)Todorov, Erez, and Tassa]{mujoco}
Emanuel Todorov, Tom Erez, and Yuval Tassa.
\newblock Mujoco: A physics engine for model-based control.
\newblock In \emph{IEEE/RSJ International Conference on Intelligent Robots and
  Systems (IROS)}, pages 5026--5033. IEEE, 2012.

\bibitem[Plappert et~al.(2018)Plappert, Andrychowicz, Ray, McGrew, Baker,
  Powell, Schneider, Tobin, Chociej, Welinder, et~al.]{fetch}
Matthias Plappert, Marcin Andrychowicz, Alex Ray, Bob McGrew, Bowen Baker,
  Glenn Powell, Jonas Schneider, Josh Tobin, Maciek Chociej, Peter Welinder,
  et~al.
\newblock Multi-goal reinforcement learning: Challenging robotics environments
  and request for research.
\newblock \emph{arXiv preprint arXiv:1802.09464}, 2018.

\bibitem[Fern{\'a}ndez and Veloso(2006)]{fernandez2006probabilistic}
Fernando Fern{\'a}ndez and Manuela Veloso.
\newblock Probabilistic policy reuse in a reinforcement learning agent.
\newblock In \emph{Proceedings of the Fifth International Joint Conference on
  Autonomous Agents and Multiagent Systems}, pages 720--727. ACM, 2006.

\bibitem[Kingma and Ba(2014)]{adam}
Diederik~P Kingma and Jimmy Ba.
\newblock Adam: A method for stochastic optimization.
\newblock \emph{arXiv preprint arXiv:1412.6980}, 2014.

\end{thebibliography}

\section{Appendix}

\subsection{Initial Policies}
\label{sec:initial-policies-details}

Here we describe in detail the initial policies used for our experiments.

\subsubsection{\texttt{DiscreteMPCPush}}

Suppose we have a learned or known transition model $Pr(s' | s, a)$ that can be queried to predict the state trajectories and rewards that may result from a sequence of actions taken from an initial state. In Model-Predictive Control (MPC), we use this transition model to select each action taken by the policy $\pi$. More specifically, given the current state $s_t$, an MPC policy will internally consider multiple sequences of actions $a_t, a_{t+1}, ..., a_{h}$ and compute the expected rewards accrued for each sequence. The first action in the best sequence is then the output of $\pi(s_t)$. To design an MPC policy, we must therefore specify the model and a procedure for selecting possible action sequences.

In high-dimensional tasks with long horizons, sparse rewards, and continuous states and actions, MPC is intractable without an efficient mechanism for selecting action sequences. Here we opt to discretize the action space as a means to simplify the search. In particular, rather than consider the infinite number of possible gripper movements, we consider only six, one per cardinal direction. We can then use a discrete graph search to explore possible action sequences.

We develop a discrete MPC policy for the \texttt{Push} task. The model is a perfect copy of the environment (i.e. a separate instance of MuJoCo). We further improve the policy by introducing an informative heuristic to guide the discrete search. The heuristic is a tuple $(d_1, d_2)$ where $d_1$ is the distance between the object and the target location and $d_2$ is the distance between the gripper and the ``push location.'' The push location is meant to be the desired position of the gripper for pushing the block to the target; it is approximated by extending the vector difference between the object and target location by a small amount corresponding to the radius of a sphere circumscribed around the object. The second entry $d_2$ of the heuristic is only used to break ties when the first entry $d_1$ matches. We use this heuristic to perform a best-first search with 10 node expansions per environment step. At the end of the search, we find the node with the best heuristic value and take the corresponding first action. (If the root has the best heuristic value, we take a noop action.)

\subsubsection{\texttt{ReactivePush}}

Our second policy is designed for pushing an object to a target location. While this policy works nearly perfectly in the original \texttt{Push} task, its performance drops dramatically when the sliding friction on the block is reduced as in the \texttt{SlipperyPush} task. Given an input state, the policy checks the following conditional statements in order until one holds and proceeds accordingly.

\begin{enumerate}
    \item If the object is already at the target location, do nothing.
    \item If the block is between the gripper and the target location, move the gripper towards the target location.
    \item If the gripper is above the push location (see definition in \texttt{DiscreteMPCPush}), move the gripper down to prepare to push.
    \item Move the gripper to above the push location.
\end{enumerate}

\noindent To determine whether the object or gripper is ``at'' a location, we measure the distance and check if it is below a global threshold. The other key hyperparameter is a gain that determines how far the gripper moves at each time step. We manually tuned this gain to achieve near optimal performance on the original \texttt{Push} task.

\subsubsection{\texttt{ReactivePickAndPlace}}

Our third policy is designed to pick up a cube and bring it to a target location on or above the table. Given an input state, the policy checks the following conditional statements in order until one holds and proceeds accordingly.

\begin{enumerate}
    \item If the object is already at the target location, do nothing.
    \item If the gripper is grasping the object, move towards the target location.
    \item If the object is between the gripper fingers (but not grasped), close the gripper.
    \item If the gripper is above the object
    \item \quad ... and the gripper is closed, open the gripper.
    \item \quad ... and the gripper is open, move the gripper down.
    \item Move the gripper towards the location above the object.
\end{enumerate}

\noindent To determine whether the gripper is grasping the object, we check that the object location is between the two fingers and that the fingers are not more than the block width apart. We again use the distance threshold and gain hyperparameters described above.

\subsubsection{\texttt{ReactiveHook}}

Our fourth policy is designed to pick up a hook, move it behind and to the right of an object, and push and pull the object towards a target location. The policy works nearly perfectly when the object is a cube, the table is clear of obstacles, and the observations are noise-free (see Figure\ref{fig:hook-intro}a). However, the policy performance drops substantially when transferred to the \texttt{NoisyHook} and \texttt{ComplexHook} tasks. Given an input state, the policy checks the following conditional statements in order until one holds and proceeds accordingly.

\begin{enumerate}
    \item If the object is already at the target location, do nothing.
    \item If the hook is not grasped and lifted above the table, grasp and lift the hook.
    \item If the hook is not beyond and to the right of the object, move forward or rightward accordingly.
    \item Move the gripper following the vector difference between the object and the target location.
\end{enumerate}

The grasp position is fixed so that the robot always attempts to pick up the same part of the hook (near the bottom). In addition to the global threshold and gain hyperparameters, we use knowledge of the length and width of the hook to determine gripper movements as a function of desired hook movements.

\subsubsection{\texttt{CachedPETS}}

Our final policy uses a model-predictive controller (MPC) with a learned transition model. We take the recently proposed Probabilistic Ensembles with Trajectory Sampling (PETS) as our method for model-based reinforcement learning \cite{chua2018deep}. PETS learns a transition model in the form of an ensemble of probabilistic neural networks. During planning, a sequence of actions is sampled with reference to previously high-reward action sequences using the cross-entropy method. To predict the subsequent states and rewards using the learned transition model, a finite collection of particles are propagated forward in time. The action that leads to the highest expected reward is selected, and planning repeats after each environment step. We use the PETS implementation made available by \citet{chua2018deep} without modification.

MPC methods are generally much slower than model-free counterparts. Indeed, we found PETS alone to be intractably slow as an initial policy for RPL. We therefore create a ``cached'' version that stores the action produced by PETS for 500 input states. The number 500 was selected based on a small performance analysis (see later in the appendix). We select these 500 input states by sampling trajectories from the environment on-policy. At test time, given a new state, we find the nearest state in the cache (as measured by Euclidean distance) and take the corresponding action. Though quite simple and coarse as an approximation of the full MPC, the final \texttt{CachedPETS} controller performs only slightly worse than the original PETS (see Figure \ref{fig:empirical-results-group2}c) with a 2-3 order of magnitude speed up.

\begin{figure}[t!]
\centering
  \includegraphics[width=0.5\textwidth]{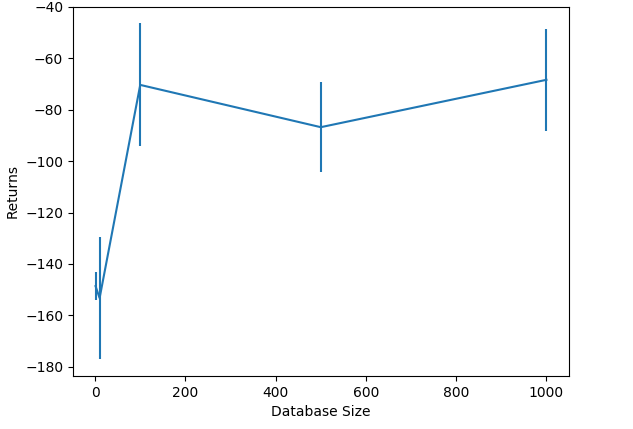}
  \caption{To select the size of the cache used for the \texttt{CachedPETS} controller, we plot performance on the \texttt{MBRLPusher} task for sizes ranging from 1 to 1000. We run 25 trials for each size and plot the mean and standard deviation here. Based on this analysis, we selected a database of size 500.}
  \label{fig:database-analysis}
\end{figure}

\subsection{Model Hyperparameters}
All experiments in this paper use the following hyperparameters, which are taken from \cite{fetch}.
\begin{itemize}
    \item Actor and critic networks: $3$ layers with $256$ units each and ReLU non-linearities
    \item Adam optimizer~\citep{adam} with $1\cdot10^{-3}$ for training both actor and critic
    \item Buffer size: $10^6$ transitions
    \item Polyak-averaging coefficient: $0.95$
    \item Action L2 norm coefficient: $1.0$
    \item Observation clipping: $[-200, 200]$
    \item Batch size: $256$
    \item Cycles per epoch: $50$
    \item Batches per cycle: $40$
    \item Test rollouts per epoch: $50$
    \item Probability of random actions: $0.3$
    \item Scale of additive Gaussian noise: $0.2$ ($0.1$ for hooks)
    \item Probability of HER experience replay: $0.8$
    \item Normalized clipping: $[-5, 5]$
\end{itemize}

\noindent For the \texttt{Push} and \texttt{PickAndPlace} experiments, we use 19 MPI workers with a rollout batch size of 2 to match the previous work. For the \texttt{Hook} experiments, we use 1 MPI worker and a rollout batch size of 4 to save on compute resources. We determined the "Expert Explore" baseline hyperparameters $\epsilon=0.6$ and $\alpha=0.8$ with a small grid search. For all experiments, we used a burn-in threshold of $\beta = 1.0$. We did not optimize this hyperparameter and believe RPL's performance could be further improved in doing so.

\end{document}